\documentclass[runningheads]{llncs}
\usepackage[T1]{fontenc}
\usepackage{graphicx}
\usepackage{hyperref}
\usepackage{color}

\usepackage{siunitx}
\usepackage{xcolor}
\usepackage{amsmath}
\usepackage{amssymb}
\usepackage{booktabs}
\usepackage{multirow}

\usepackage[misc]{ifsym}

\begin{document}
\title{Adapting the Mean Teacher for keypoint-based lung registration under geometric domain shifts}
\titlerunning{Adapting the Mean Teacher for registration under geometric domain shifts}
%
\author{Alexander Bigalke\textsuperscript{(\Letter)}\orcidID{0000-0001-7824-5735} \and
Lasse Hansen\orcidID{0000-0003-3963-7052} \and
Mattias P. Heinrich\orcidID{0000-0002-7489-1972}}

\authorrunning{A. Bigalke et al.}
%
\institute{Institute of Medical Informatics, University of L\"ubeck, Germany \email{\{alexander.bigalke,l.hansen,mattias.heinrich\}@uni-luebeck.de}}
%
\maketitle              
\begin{abstract}
Recent deep learning-based methods for medical image registration achieve results that are competitive with conventional optimization algorithms at reduced run times.
However, deep neural networks generally require plenty of labeled training data and are vulnerable to domain shifts between training and test data.
While typical intensity shifts can be mitigated by keypoint-based registration, these methods still suffer from geometric domain shifts, for instance, due to different fields of view.
As a remedy, in this work, we present a novel approach to geometric domain adaptation for image registration, adapting a model from a labeled source to an unlabeled target domain.
We build on a keypoint-based registration model, combining graph convolutions for geometric feature learning with loopy belief optimization, and propose to reduce the domain shift through self-ensembling.
To this end, we embed the model into the Mean Teacher paradigm.
We extend the Mean Teacher to this context by 1) adapting the stochastic augmentation scheme and 2) combining learned feature extraction with differentiable optimization.
This enables us to guide the learning process in the unlabeled target domain by enforcing consistent predictions of the learning student and the temporally averaged teacher model.
We evaluate the method for exhale-to-inhale lung CT registration under two challenging adaptation scenarios (DIR-Lab 4D CT to COPD, COPD to Learn2Reg).
Our method consistently improves on the baseline model by \SI{50}{\%}/\SI{47}{\%} while even matching the accuracy of models trained on target data.
Source code is available at \url{https://github.com/multimodallearning/registration-da-mean-teacher}.

\end{abstract}

\section{Introduction}
Image registration is a fundamental task in medical image analysis, for instance required for multi-modal data fusion or patient monitoring over time.
For a long time, the state of the art for image registration was dominated by conventional optimization methods \cite{sotiras2013deformable}, whose high accuracy comes at the cost of high run times. 
In recent years, learning-based methods---driven by deep neural networks---achieved competitive performances \cite{haskins2020deep}.
These methods substantially reduce inference times, but they involve two other significant drawbacks.
First, high performance is strongly dependent on the availability of labeled training data, which are costly to collect.
Second, deep neural networks often generalize poorly to shifted domains. 
Once trained in a labeled source domain, the models are likely to suffer a performance drop when deployed on data from a shifted target domain.
While shifts in intensity distributions---typical for medical imaging---can be mitigated by keypoint-based registration \cite{hansen2021deep}, such methods still suffer from geometric domain shifts, for instance, due to varying fields of view under different imaging protocols.
Fine-tuning or re-training on data from the shifted domain could alleviate the performance drop but is often impractical due to high labeling costs.
Alternatively, domain adaptation \cite{wang2018deep} is a promising technique to adapt a model from a labeled source to an unlabeled target domain.
While extensively explored for medical classification and segmentation tasks \cite{guan2021domain}, domain adaptation for image registration has rarely been studied in the literature \cite{kruse2021multi,mahapatra2020training} and will be the focus of this work.

Existing works on domain adaptive registration rely on two different concepts. Mahapatra et al.~\cite{mahapatra2020training} increase the invariance of a generative registration model to the type of input images by encoding the images to the latent space of an autoencoder.
In a different approach, Kruse et al.~\cite{kruse2021multi} adapt the concept of maximum classifier discrepancy \cite{saito2018maximum} to multi-modal registration.
However, this requires the quantization of displacement fields and the use of a classification architecture instead of state-of-the-art registration models.
Generally, domain adaptive registration is challenging because concepts established for other tasks are often unsuitable for registration.
The mainstream approach of domain-invariant feature learning through adversarial learning \cite{ganin2015unsupervised,tzeng2017adversarial} or reconstruction \cite{bousmalis2016domain,ghifary2016deep}, for instance, was primarily used in classification tasks.
In consequence, the methods focus on the alignment of global feature vectors.
This is insufficient for registration, which highly depends on the identification of local correspondences.
Alternatively, Tsai et al.~\cite{tsai2018learning} proposed domain adaptation in the output space, where the distributions of predictions in source and target domain are aligned through adversarial learning.
This concept was successfully applied to semantic segmentation \cite{tsai2018learning} and human pose estimation \cite{yang20183d}.
However, the raw displacement fields output by registration models are less structured than segmentation masks or human poses such that aligning their distributions might be ineffective.
Instead of distribution matching, self-ensembling \cite{french2018selfensembling} addresses the domain shift by imposing consistency constraints in the output space.
The method is based on the framework of the Mean Teacher \cite{tarvainen2017mean}, comprising a learning student model and a so-called teacher model, which represents a temporal ensemble with its weights corresponding to the exponential moving average (EMA) of the student model.
Supervision on unlabeled target data is then provided by enforcing consistent predictions of student and teacher model.
Beyond classification \cite{french2018selfensembling}, this concept has successfully been adapted to diverse tasks, including medical image segmentation \cite{perone2019unsupervised,yu2019uncertainty} and clinical human pose estimation \cite{srivastav2021unsupervised}.

\begin{figure}[t]
\centering
\includegraphics[width=\textwidth]{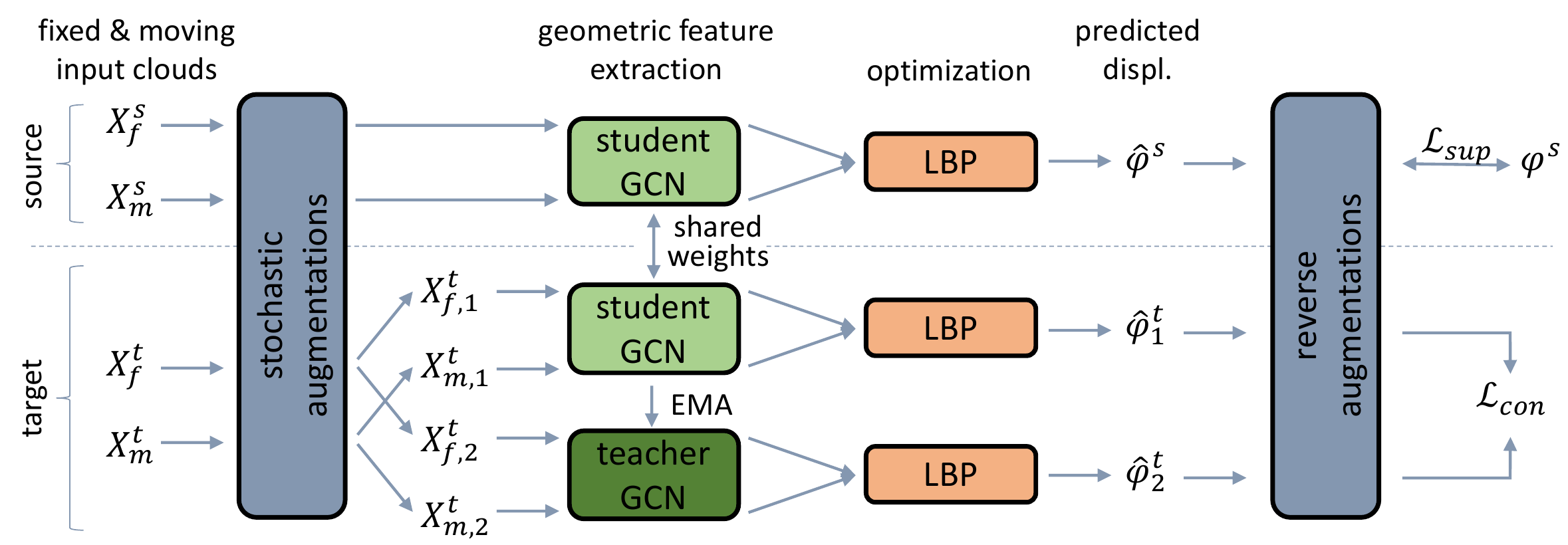}
\caption{Overview of our proposed self-ensembling framework for domain adaptive keypoint-based image registration.}
\label{fig:overview}
\end{figure}

\textbf{Contributions}. In this work, we extend the concept of self-ensembling to geometric domain adaptation for image registration, embedding a keypoint-based registration model into the Mean Teacher paradigm (see Fig.~\ref{fig:overview}).
Our framework is built upon the keypoint-based method by Hansen et al.~\cite{hansen2021deep}, which aligns two point clouds (extracted from the input scan pair) by combining a Graph Convolutional Network (GCN) for learned geometric feature extraction with loopy belief propagation (LBP) for alignment.
To incorporate the method into the Mean Teacher framework, we present two crucial modifications of the standard Mean Teacher.
First, we adapt the stochastic augmentation scheme to the specific characteristics of the registration task by incorporating inverse geometric transformations.
Second, we present the first Mean Teacher that combines learned feature extraction (GCN) with differentiable optimization (LBP).
Notably, the differentiability of LBP enables us to impose the consistency constraint between student and teacher model on the final predicted displacement fields after LBP.
That way, the adaptation of the GCN can benefit from the regularizing effect of LBP.
Overall, our method offers a simple and robust adaptation procedure.
It involves only few hyper-parameters and does not rely on intricate adversarial optimization as previous works.
We experimentally evaluate our method for exhale-to-inhale lung CT registration, considering two different geometric domain shifts (varying field of view and breathing type).
Results demonstrate substantial improvements compared to the baseline model and even competitive performance with fully-supervised models.

\section{Methods}
\subsection{Problem Statement}
Given fixed and moving input scans $\boldsymbol{F},\boldsymbol{M}\in\mathbb{R}^{H\times W\times D}$, the goal of registration is to predict a displacement field $\boldsymbol{\varphi}$ that spatially aligns $\boldsymbol{M}$ to $\boldsymbol{F}$ within a given region of interest.
We address this task in the classical domain adaptation setting, where training data comprises a labeled source dataset $\mathcal{S}$ and an unlabeled target dataset $\mathcal{T}$.
$\mathcal{S}$ consists of triplets $(\boldsymbol{F}^s,\boldsymbol{M}^s,\boldsymbol{\varphi}^s)$ of scan pairs $(\boldsymbol{F}^s,\boldsymbol{M}^s)$ with corresponding ground truth displacement fields $\boldsymbol{\varphi}^s$.
$\mathcal{T}$ contains scan pairs $(\boldsymbol{F}^t,\boldsymbol{M}^t)$ without any ground truth.
Given the training data, we aim to learn a function $f$ with parameters $\boldsymbol{\theta}$ that predicts displacement fields $\boldsymbol{\hat{\varphi}}=f(\boldsymbol{F},\boldsymbol{M};\boldsymbol{\theta})$ and achieves the best possible performance on the target domain.
In the following, we first introduce our baseline model $f$ for standard supervised learning on source data (Sec.~\ref{sec:baseline_model}) and subsequently present our self-ensembling framework for domain adaptation (Sec.~\ref{sec:mean_teacher}).

\subsection{Baseline Model}
\label{sec:baseline_model}
Most learning-based methods perform registration based on intensities in dense voxel space.
Instead, we formulate registration as the pure geometric alignment of two point clouds $\boldsymbol{X}_f\in\mathbb{R}^{N_f\times 3},\boldsymbol{X}_m\in\mathbb{R}^{N_m\times 3}$, representing distinct keypoints of the input scans.
The underlying motivation is that this reduces the vulnerability to intensity shifts between source and target domain, enabling us to focus on remaining geometric shifts.
The general efficacy of learning-based registration of point clouds in a fully-supervised setting has previously been demonstrated by the authors of \cite{hansen2021deep}.
We adopt their model as our baseline and briefly summarize its major components.

First, fixed and moving point clouds $\boldsymbol{X}_f$ and $\boldsymbol{X}_m$ are extracted from the input scans using the Foerstner algorithm and non-maximum suppression \cite{heinrich2015estimating}.
Second, a GCN $g$ extracts point-wise geometric features $g(\boldsymbol{X}_f;\boldsymbol{\theta}_g)$, $g(\boldsymbol{X}_m;\boldsymbol{\theta}_g)$ from the raw coordinates of both clouds.
The GCN is based on the edge convolution from \cite{wang2018deep}, operating on the knn-graph of the clouds to account for neighborhood relations.
Third, the extracted features are used to guide the inference of final displacement vectors with LBP.
Specifically, correspondence probabilities between fixed keypoints and candidate sets from the moving cloud are computed via LBP by jointly minimizing a data cost and a regularization cost.
While the latter enforces smoothness of the predicted displacement field, the data cost is defined as the distance between geometric features and induces high correspondence probabilities between points with similar features.
Thus, effective feature extraction with the GCN is crucial for the assignment of accurate correspondences.
Finally, displacement vectors are inferred by integrating the probabilities over the corresponding displacements between fixed and candidate points, allowing for more accurate displacement vectors than hard assignments.
We formally summarize the combination of GCN and LBP as
\begin{equation}
    \hat{\boldsymbol{\varphi}}=LBP(g(\boldsymbol{X}_f;\boldsymbol{\theta}_g), g(\boldsymbol{X}_m;\boldsymbol{\theta}_g))=:f(\boldsymbol{X}_f,\boldsymbol{X}_m;\boldsymbol{\theta}),\quad \hat{\boldsymbol{\varphi}}\in\mathbb{R}^{N_f\times 3}
\end{equation}
and refer the reader to \cite{hansen2021deep} for more details.
As such, $f$ is fully differentiable, enabling end-to-end learning of parameters $\boldsymbol{\theta}$ by minimizing the supervised loss
\begin{equation}
    \label{eq:loss-sup}
    \mathcal{L}_{sup}=\|\hat{\boldsymbol{\varphi}}-\boldsymbol{\varphi}\|_1
\end{equation}

\subsection{Domain-adaptive Registration with the Mean Teacher}
\label{sec:mean_teacher}
To adapt the baseline model to a shifted target domain, we propose a novel self-ensembling framework, embedding the model into the Mean Teacher paradigm \cite{french2018selfensembling,tarvainen2017mean}.
An overview of the method is shown in Fig.~\ref{fig:overview}.
The framework extends the baseline model, and now includes two GCNs for feature extraction, namely a student GCN with weights $\boldsymbol{\theta}$ and a teacher GCN with weights $\boldsymbol{\theta}'$.
The student network is optimized via gradient descent, whereas the weights of the teacher are the exponential moving average of the student's weights, updated as
\begin{equation}
    \label{eq:teacher_update}
    \boldsymbol{\theta}'_i=\alpha\boldsymbol{\theta}'_{i-1}+(1-\alpha)\boldsymbol{\theta}_i
\end{equation}
at iteration $i$ with the momentum $\alpha$.
The student is optimized by minimizing the joint loss function 
\begin{equation}
    \label{eq:total-loss}
    \mathcal{L}(\boldsymbol{\theta};\boldsymbol{\theta}',\mathcal{S},\mathcal{T})=\mathcal{L}_{sup}(\boldsymbol{\theta};\mathcal{S})+\lambda(t)\mathcal{L}_{con}(\boldsymbol{\theta};\boldsymbol{\theta}',\mathcal{T})
\end{equation}
composed of the supervised loss $\mathcal{L}_{sup}$ (cf.~Eq.~\ref{eq:loss-sup}) on labeled source data and a consistency loss $\mathcal{L}_{con}$ on unlabeled target data, weighted by a time-dependent factor $\lambda(t)$.
The consistency loss penalizes different predictions by student and teacher model.
Implementing $\mathcal{L}_{con}$ for the given registration problem requires two subtle but decisive adaptations of the standard Mean Teacher for classification.

First, the standard Mean Teacher imposes the consistency constraint at the output of the learning network, which usually coincides with the final prediction.
In our case, however, the output of the learning GCN is an intermediate representation, and we observed that applying consistency constraints at this level hampers the learning process.
Instead, we propose to align predicted displacement vectors after LBP.
That way, the adaptation process can benefit from the regularizing effect of LBP.

Second, an important component of the Mean Teacher is the application of different stochastic augmentations to the input of teacher and student streams.
Unlike classification, however, registration requires the augmentation of pairs of inputs instead of single inputs.
Moreover, the associated displacement fields in the output space are not invariant to the input transformation (as is the case for classification), but they are transformed together with the input.
For $\mathcal{L}_{con}$ to be meaningful, transformations applied at the input level need to be reversed at the output level such that predictions in teacher and student streams are aligned.
To ensure reversibility at the output level, we sample one transformation for each stream and apply it to both fixed and moving cloud, yielding two augmented pairs ($\boldsymbol{X}_{f,1}^t,\boldsymbol{X}_{m,1}^t)$,($\boldsymbol{X}_{f,2}^t,\boldsymbol{X}_{m,2}^t)$.
In practice, we transform point clouds by random rotation, translation and scaling.
The reverse transformation of the displacement vectors is then given by inverse scaling and rotation (displacement vectors are invariant to the synchronous translation of both inputs).
Denoting reverse augmentations as $\mathrm{aug}^{-1}(.)$, we formalize the consistency loss as 
\begin{equation}
    \label{eq:cons_loss}
    \mathcal{L}_{con}=\left\|\mathrm{aug}_{1}^{-1}(f(\boldsymbol{X}_{f,1}^t,\boldsymbol{X}_{m,1}^t;\boldsymbol{\theta}))-\mathrm{aug}_{2}^{-1}(f(\boldsymbol{X}_{f,2}^t,\boldsymbol{X}_{m,2}^t;\boldsymbol{\theta}'))\right\|_1
\end{equation}
At early epochs, the weights of the teacher model are still close to the random initialization, inducing noisy gradients from $\mathcal{L}_{con}$.
Therefore, the weighting factor $\lambda(t)=\lambda_0\cdot\exp({-5(1-\min(t/T,1)^2}))$ depends on the current epoch $t$ and steadily increases from 0 to $\lambda_0$ during the first $T$ epochs, as suggested by \cite{tarvainen2017mean}.

\section{Experiments}
\subsection{Experimental Setup}
\subsubsection{Datasets and Pre-processing.}
We evaluate our method for exhale-to-inhale lung CT registration under two adaptation scenarios using three public datasets.
First, we consider the DIR-Lab COPD dataset \cite{castillo2013reference} as source and the Learn2Reg (L2R) Task 2 dataset \cite{hering_alessa_2020_3835682} as target domain.
The domain shift consists in exhale scans from the target domain exhibiting a cropped field of view such that upper and lower parts of the lungs are partially cut off.
Our training data comprise 10 labeled scan pairs from COPD and 12 unlabeled pairs from the train/val split of L2R.
Evaluation is performed on the official test split of L2R (10 scan pairs).
Second, we perform adaptation from the DIR-Lab 4D CT dataset \cite{castillo2009framework} as the source to the COPD dataset as the target domain.
While scans from 4D CT were acquired from patients with shallow resting breathing, scan pairs from COPD show actively forced full inhalation and exhalation.
Thus, the domain gap consists in the breathing type, yielding deformations with different characteristics.
Here, evaluation is performed via 5-fold cross-validation on the 10 scans from COPD.
For each fold, training data include the 10 labeled scan pairs from 4D CT and 8 unlabeled scan pairs from COPD, and we evaluate the trained model on the remaining 2 scan pairs from COPD.
In both adaptation scenarios, labels for source data are available in the form of landmark correspondences.
To supervise predicted displacement vectors for the keypoints in the fixed cloud, we interpolate displacements of the landmarks to the entire volume and grid-sample at the keypoint locations.
Scans from all domains are pre-processed in an identical way.
We resample inhale scans to $1.75\times1.25\times\SI{1.75}{mm}$ and exhale scans to $1.75\times1.00\times\SI{1.25}{mm}$ to compensate for a different volume scaling during inspiration and expiration.
Subsequently, we crop fixed-size regions of interest with $192\times192\times\SI{208}{vx}$ around the centre of automatically generated lung masks.

\subsubsection{Implementation Details.}
We implement our framework in PyTorch and optimize parameters of the student GCN with the Adam optimizer.
Batches are composed of 4 source and 4 target samples.
Training is performed for 100 epochs with a constant learning rate of 0.01.
Hyper-parameters of the baseline model (GCN architecture and LBP parameters) are adopted from \cite{hansen2021deep}.
Hyper-parameters of the self-ensembling framework are determined via cross-validation on the training samples from Learn2Reg and set to $\lambda_0=1$, $\alpha=0.98$, and $T=20$ epochs.
Stochastic augmentations include random rotations around all axes by angles from $[-10^{\circ},10^{\circ}]$, random scaling by a global factor from $[0.9,1.1]$, and random translation by a vector from $[-0.1, 0.1]^3$.

\subsubsection{Baseline Methods.}
1) As a lower bound (\textit{source-only}), we train the baseline model on source data without any adaptation techniques.
2) As an upper bound (\textit{target-only}), we train the baseline model on labeled training data from the target domain.
3) As an alternative domain adaptation technique, we experimented with DANN \cite{ganin2015unsupervised}.
Specifically, we used max pooling to condense geometric features into a global feature vector and aligned the source and target distributions of these global features by adversarial learning.
As expected, this approach turned out to be unsuitable for registration and lead to divergence.
4) As an intensity-based baseline method, we use the recent Voxelmorph++ (\textit{VM++}) \cite{heinrich2022voxelmorph}, which combines an extension of Voxelmorph with instance optimization through Adam.
We train the model once on source and once on target data.

\subsubsection{Metrics.}
We interpolate predicted displacement vectors at sparse keypoints to the entire volume and report the mean target registration error (TRE) at available landmark correspondences.
We also inspected the regularity of the interpolated deformation fields.
However, due to the regularizing effect of LBP, all methods predict smooth deformation fields (percentages of folding voxels $|J_{\varphi}|_{<0}<10^{-4}$) such that a quantitative comparison does not provide additional insights.

\subsection{Results}
Quantitative results of our experiments are shown in Tab.~\ref{results_tab} and reveal consistent findings under both adaptation scenarios.
\begin{table}[t]
\begin{minipage}{0.35\textwidth}
\caption{Results for both adaptation scenarios, reported as TRE in mm.}
\label{results_tab}
\begin{tabular}{lcc}
\toprule
\multirow{2}{*}{Method} &  COPD & \multicolumn{1}{l}{4D CT}\\
& $\rightarrow$L2R & $\rightarrow$COPD\\
\midrule
initial & 10.24 & 11.99\\
VM++ source &  4.34 & 3.55\\
VM++ target & 3.09 & 2.46\\ 
source-only &  4.96 & 4.02\\
target-only & 2.61 & 2.26\\
ours & 2.64 & 2.01\\
ours + Adam & 2.38 & 1.93\\
\bottomrule
\end{tabular}
\end{minipage}
\hfill
\begin{minipage}{0.55\textwidth}
\centering
\includegraphics[width=\textwidth]{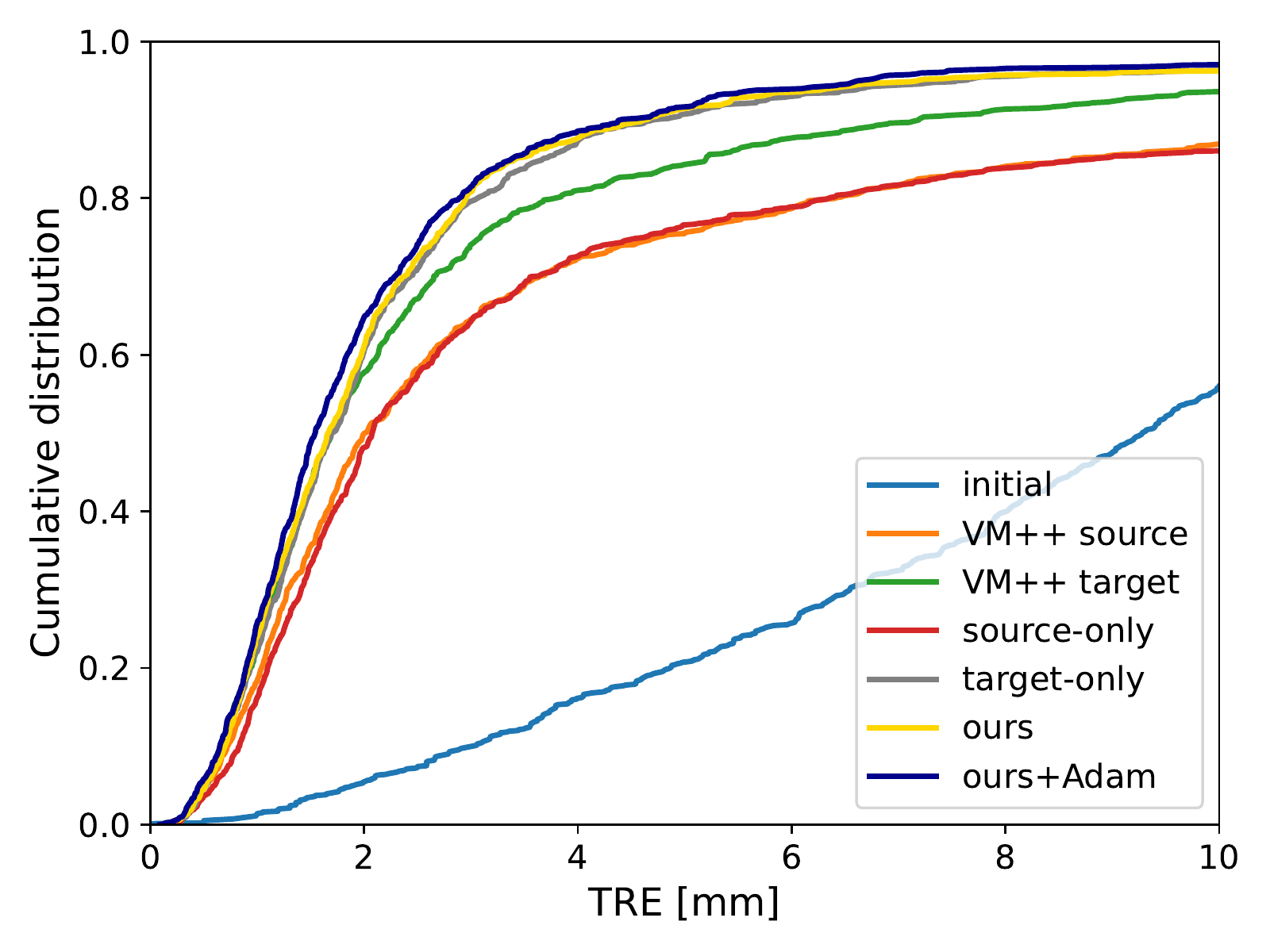}
\renewcommand\tablename{Fig.}
\vspace{-0.4cm}
\caption{Cumulative target registration errors for adaptation from COPD to L2R.}
\label{fig:cum_tre}
\end{minipage}
\end{table}
First, both our source-only model and the VM++ source model perform substantially worse than their respective counterparts trained on target data, increasing the TRE by 40 to \SI{90}{\%}.
This demonstrates that both considered domain shifts pose severe problems for intensity- and keypoint-based methods and need to be addressed by effective adaptation methods.
Second, and most importantly, our proposed method effectively adapts the baseline model to the target domain.
Specifically, it achieves mean TREs of \SI{2.64}{mm} and \SI{2.01}{mm}, respectively, thus improving on the source-only model by \SI{47}{\%}/\SI{50}{\%} while matching or even surpassing the performance of the target-only model.
This demonstrates that our self-ensembling framework effectively leverages unlabeled target data to close the domain gap.
Third, we investigate how far we can push the accuracy of our method by fine-tuning the displacement fields with MIND-based instance optimization with Adam \cite{siebert2021fast}.
This further reduces the TRE to \SI{2.38}{mm} and \SI{1.93}{mm}.
A detailed comparison of all discussed methods for COPD$\rightarrow$L2R is shown in Fig.~\ref{fig:cum_tre}.
Finally, it is notable that our method compares favorably to the state-of-the-art LapIRN \cite{mok2021conditional}, the winner of the recent Learn2Reg challenge, which is slightly superior on the L2R test set (TRE=\SI{1.98}{mm}) but clearly inferior to our method on COPD (TRE=\SI{3.83}{mm}).

\begin{figure}[t]
\renewcommand{\thefigure}{3}
\centering
\includegraphics[width=0.9\textwidth]{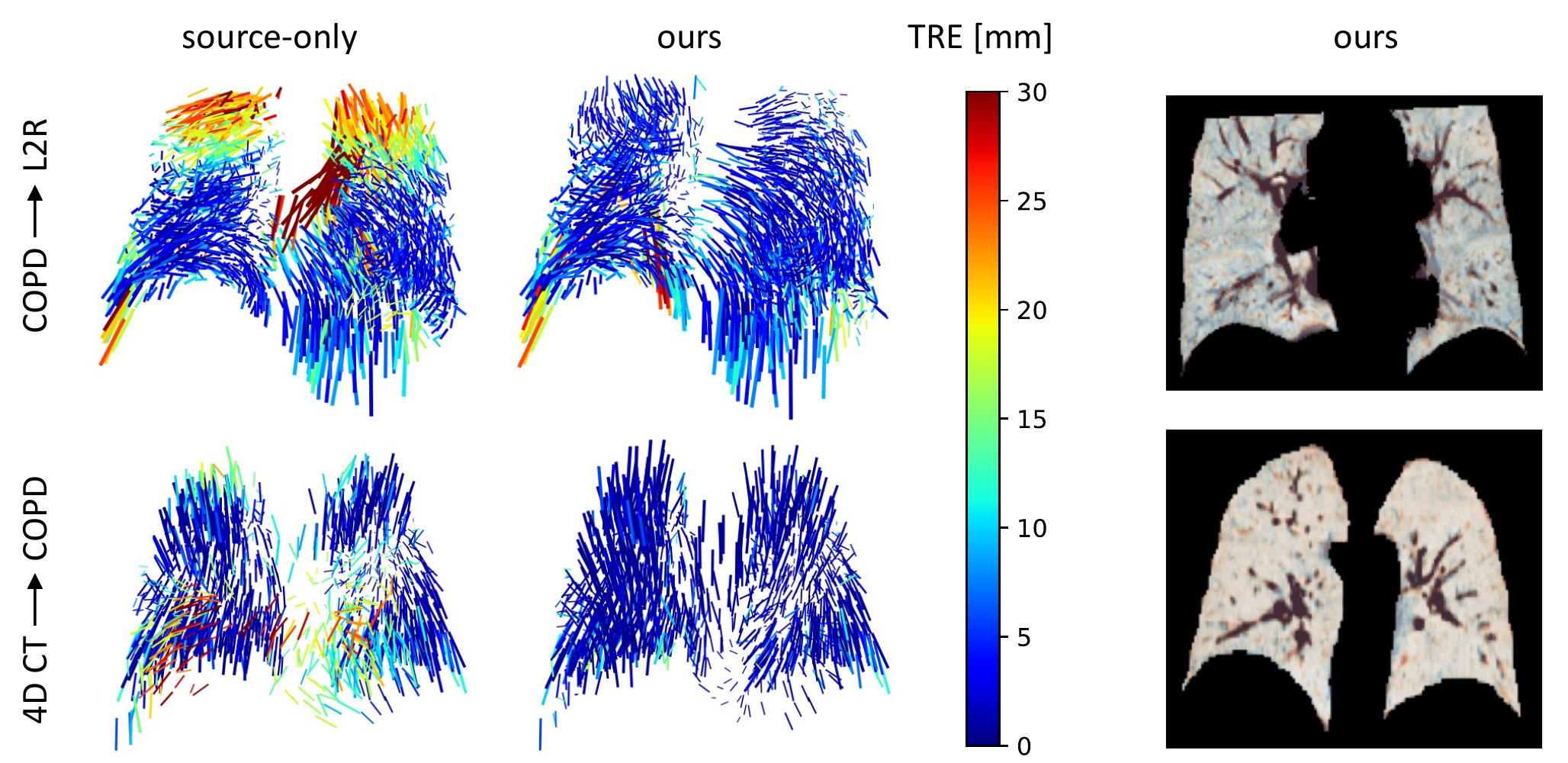}
\caption{Qualitative results for one sample case from each adaptation scenario. The first two columns show predicted displacement vectors by the source-only model and by our model. The linewidth is proportional to the distance of displacements, and colors encode the TRE (clamped to \SI{30}{mm}). The last column shows overlaid CT slices after registration by our method. Inhale and exhale scans are shown in orange and blue shades, respectively, adding up to grayscale in case of alignment.}
\label{fig:qual_results}
\end{figure}

Qualitative results are presented in Fig.~\ref{fig:qual_results}.
The first two columns visualize the effect of the studied domain gaps.
For COPD$\rightarrow$L2R, errors of the source-only model primarily occur in the superior part of the lung, which is partially outside the scanning region in exhale scans of the target domain while fully visible on source data.
For 4DCT$\rightarrow$COPD, the source-only model mainly fails in the anterior region of the lung, which is strongly deformed by full inspiration-expiration (target domain) but relatively static during shallow breathing (source domain).
Our method substantially reduces these errors, highlighting the efficient adaptation to the target domain.
Finally, the CT overlays (last column) show the largely accurate alignment of inner lung structures by our method.

\section{Conclusion}
In this work, we addressed geometric domain adaptation in the context of image registration and proposed a novel self-ensembling framework.
Specifically, we embedded a keypoint-based registration model into the Mean Teacher paradigm and thus guided the learning process in the target domain by consistency-based supervision.
In our experiments for exhale-to-inhale registration of lung CT scans, we demonstrated that our method successfully reduced the domain gap under two challenging adaptation settings, including different breathing types and imaging protocols.
Specifically, it surpassed the baseline model by \SI{50}{\%}/\SI{47}{\%} and even matched the performance of a supervised model trained on labeled target data.
These results indicate great potential of the Mean Teacher framework for medical image registration, demonstrating its capability to improve feature learning in the absence of labels.
While our use case of keypoint-based registration under geometric shifts is rather specific, our method is flexible and can easily be adapted to diverse scenarios, including domain adaptation under intensity shifts with classical 3D CNNs and semi-supervised learning.
However, experimental evaluation under these settings is needed and will be the subject of future work.

\paragraph{Acknowledgement.} We gratefully acknowledge the financial support by the \linebreak Federal Ministry for Economic Affairs and Climate Action of Germany \linebreak (FKZ: 01MK20012B) and by the Federal Ministry for Education and Research of Germany (FKZ: 01KL2008).

%
%
%
\bibliographystyle{splncs04}
\bibliography{paper2006-bibliography}

\end{document}